\documentclass[conference]{IEEEtran}
\IEEEoverridecommandlockouts
\usepackage{cite}
\usepackage{amsmath,amssymb,amsfonts}
\usepackage{algorithmic}
\usepackage{graphicx}
\usepackage{textcomp}
\usepackage{xcolor}
\usepackage{booktabs}
\usepackage{soul} 
\usepackage{stfloats}
\usepackage{balance}
\def\BibTeX{{\rm B\kern-.05em{\sc i\kern-.025em b}\kern-.08em
    T\kern-.1667em\lower.7ex\hbox{E}\kern-.125emX}}
\usepackage{fancyhdr}
\fancypagestyle{ieeecopyright}{
  \fancyhf{} 
  \fancyfoot[C]{
    \vspace{-1.5cm} 
    \scriptsize \copyright~2026 IEEE. Personal use of this material is permitted. Permission from IEEE must be obtained for all other uses, in any current or future media, including reprinting/republishing this material for advertising or promotional purposes, creating new collective works, for resale or redistribution to servers or lists, or reuse of any copyrighted component of this work in other works.
  }
}
\begin{document}

\title{Audio emotion recognition for atypical hearing\\
}

\author{\IEEEauthorblockN{Ulysse Roussel}
\IEEEauthorblockA{\textit{EAC Team,}
\textit{STMS Lab (Sorbonne Université - Ircam - CNRS)}, Paris, France \\
roussel@ircam.fr}
}


\newcommand{\mestitres}[1]{
\hl{\textit{#1}} 
}
\newcommand{\alice}[1]{%
\marginpar{\raggedright\footnotesize
\fcolorbox{blue}{blue!15}{\parbox{1.4cm}{\textbf{A:} #1}}}}

\newcommand{\charlotte}[1]{%
\marginpar{\raggedright\footnotesize
\fcolorbox{green}{green!20}{\parbox{1.4cm}{\textbf{C:} #1}}}}

\newcommand{\isabelle}[1]{%
\marginpar{\raggedright\footnotesize
\fcolorbox{orange}{orange!20}{\parbox{1.4cm}{\textbf{I:} #1}}}}

\newcommand{\ulysse}[1]{%
\marginpar{\raggedright\footnotesize
\fcolorbox{magenta}{magenta!20}{\parbox{1.4cm}{\textbf{U:} #1}}}}

\newcommand{\yann}[1]{%
\marginpar{\raggedright\footnotesize
\fcolorbox{purple}{purple!20}{\parbox{1.4cm}{\textbf{Y:} #1}}}}

\maketitle
\thispagestyle{ieeecopyright}
\begin{abstract}

My doctoral work aims to explore Audio Emotion Recognition (AER) in the context of atypical listening. This research focuses on auditory hypersensitivity in people with autism, a phenomenon that is often difficult to evaluate and unique to each individual. Our core idea is to leverage our understanding of affect from acoustic traits, relying on the possibility of generalizing affective responses from a small amount of annotated data. As a first step, we fine-tune a large foundation model, Contrastive Language-Audio Pretraining (CLAP) using low-rank adaptation (LoRA), trained on a valence and arousal dataset of neurotypical listeners. 
\end{abstract}


\section{Introduction (key research questions)}
Every day, we navigate soundscapes without paying them much notice. When a sound strikes us, we can usually name, describe, and quantify what we feel, then ignore it. For some people this filtering breaks down entirely. Sounds that others barely register flood the senses, consume 
available energy, and make soundscapes difficult to interact with. This broader increased sensory reactivity is called hypersensitivity and is now explicitly included in the diagnostic criteria of autism spectrum disorder (ASD) (Diagnostic and Statistical Manual of Mental Disorders DSM-5). In particular, some sounds can create intense emotional reactions in individuals with ASD. 

However, despite the quantity of research in this field~\cite{Poulsen2024,Kanakri2017,Landowska2022}, much effort is still needed to understand the characteristics of auditory processing in individuals with ASD~\cite{OConnor2012, Kwong2025}.
Our goal is to bridge this gap, understand what may be disturbing, and characterize atypical perception through acoustic metrics. 

Our research is organized around three key questions:

\begin{enumerate}
    \item How can affective responses to auditory stimuli be generalized from a small number of evaluated examples?
    
    \item How to design ecological data collection protocols adapted to hypersensitive individuals?
    Protocols should reflect active listening and objectively assess emotional reaction in real-world conditions. 

    \item How to use a model generalizing the listening experience of a hypersensitive individual? 
    It is 
    important to examine how acoustic features influence affective evaluations.
\end{enumerate}

We focus  primarily on the first research question, while keeping the constraints imposed by the other two in mind.

\begin{figure}[t]
    \centering
    \includegraphics[width=8.25cm]{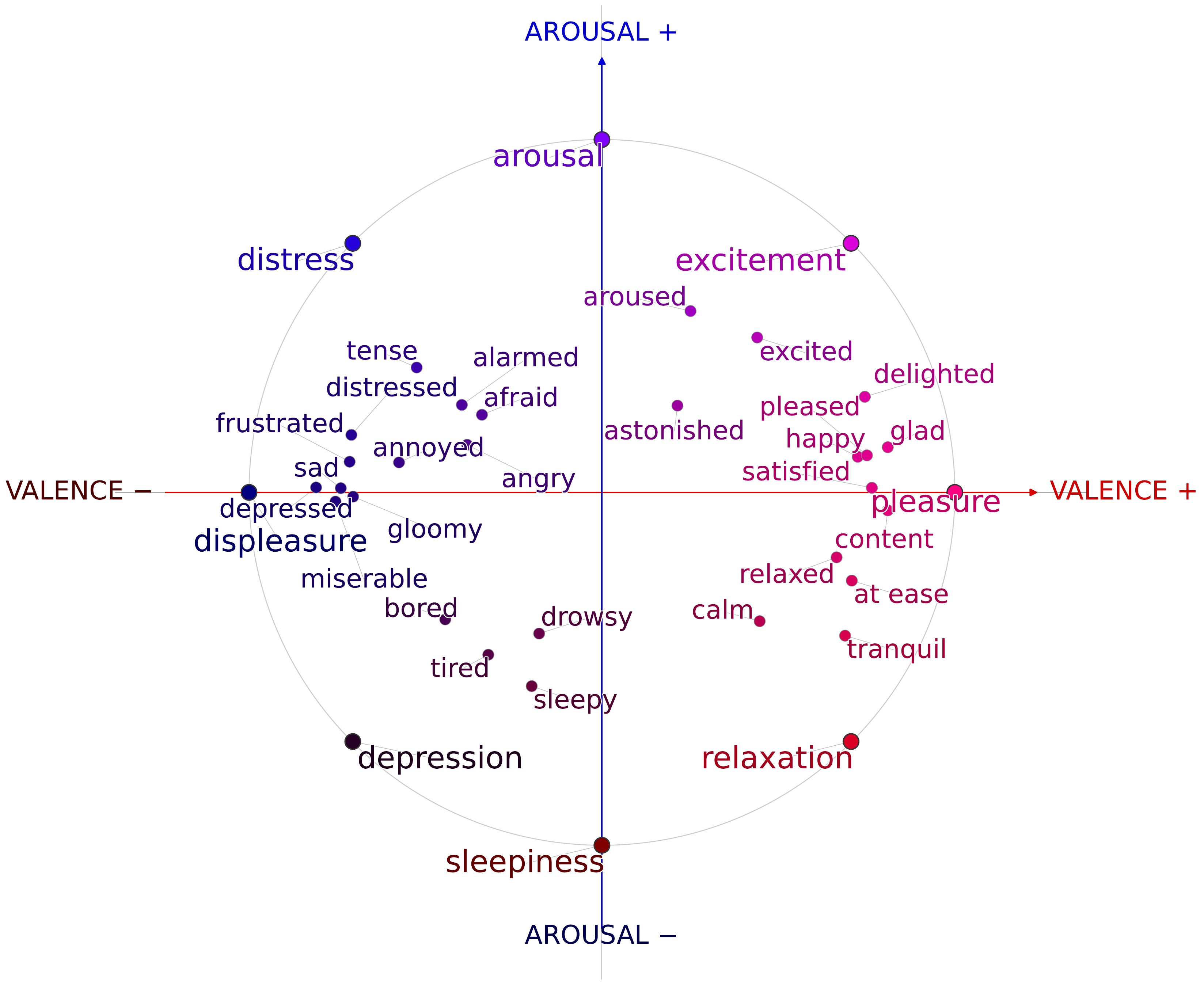}
    \caption{Russell's circumplex model of affect. The 36 emotional
    anchors used as prediction targets are distributed along the
    two dimensions: Valence (pleasure--displeasure) and Arousal
    (activated--sleepy)}
    \label{fig:russell}
\end{figure}

\section{Related work}
\label{sec:2}
\subsection{Context}

Behavioral studies report marked heterogeneity in auditory and sensory processing in ASD~\cite{Poulsen2024}, motivating a personalized approach that adapts to each listener's subjective perception from few labeled examples~\cite{fang2025metaperser}. To represent affective responses, we rely on Russell's circumplex model~\cite{Russell1980} (Fig.~\ref{fig:russell}), on which 36 subjective feelings are described along two continuous dimensions: Valence (pleasure–displeasure) and Arousal (activated–sleepy) $(V,A)$. This space is psychologically meaningful and consistent with physiological measures. These dimensions are also well-suited to machine learning approaches for predicting affective states from audio.

However, no affective audio dataset annotated by atypical listeners is currently available, and collecting such a dataset is itself a central objective of this thesis. As a first step, we validate a methodology on existing neurotypical datasets: the goal is not to model atypical listening directly, but to verify that affective responses can be generalized from limited data.

\subsection{Advantages and limitation of existing datasets}

\begin{table}[htbp]
\centering
\caption{Comparison of selected emotional audio datasets}
\label{tab:datasets}
\small
\setlength{\tabcolsep}{2pt}
\begin{tabular}{|p{0.1\textwidth}|p{0.054\textwidth}|p{0.1\textwidth}|p{0.08\textwidth}|p{0.09\textwidth}|}
\hline
\textbf{Dataset} & 
\textbf{Sounds} & 
\textbf{\% Non-mus.} & 
\textbf{n-Ratings} & 
\textbf{Annotation} \\
\hline
EmoSoundsc.    & 1{,}213 & $\sim$100\%\textsuperscript{a} & n/a & Ranking \\
IADS-E         & 935     & 74\%  & 22  & Rating  \\
IADS-2         & 167     & 69\%  & 100 & Rating  \\
AAD            & 780     & 100\% & 40  & Rating  \\
\hline
\multicolumn{5}{p{0.45\textwidth}}{\footnotesize \textsuperscript{a}Estimated; not reported in the original dataset} \\
\end{tabular}
\end{table}

\begin{enumerate}
    \item \textbf{EmoSoundscapes}~\cite{Fan2017} contains 1,213 soundscapes from Freesound.org: 600 original recordings and 613 mixes of those sources. It is annotated via pairwise ranking, which encodes relative preference. Without loudness normalization, arousal prediction may be driven by raw amplitude. The mixed sources complicate affective labeling of individual stimuli, and splits must prevent leakage across train, validation and test sets.
    \item \textbf{IADS-E}~\cite{Yang2018} expands the original International Affective Digitized Sounds (IADS) to 935 sounds, not loudness-normalized and 74\% non-musical, rated on the Self-Assessment Manikin (SAM), a discrete 1--5 scale. Each sound is rated by at least 22 participants, reducing label reliability and reproducibility.
    \item \textbf{IADS-2}~\cite{IADS2} comprises 167 non-normalized sounds rated by participants using SAM-based ratings. Despite its high annotator density (100 participants per sound), the dataset is small and contains a mix of musical, anthropomorphic and non-musical content (69\% non-musical), which limits its suitability for affective modeling of soundscapes.
    \item \textbf{AAD}~\cite{AAD} The Affective Audio Dataset (AAD) consists of 780 stimuli, exclusively non-musical, non-anthropomorphic mono sounds, normalized to $-23$\, LUFS, sourced from the BBC Sound Effects library and fixed to 6 seconds. Ratings were collected via Amazon Mechanical Turk and validated with university participants (867 raters total) using a continuous slider for fine-grained annotation. Each sound was rated by a subset of 40 participants, making AAD one of the most densely annotated non-musical affective audio datasets. 
\end{enumerate}

We thus have several relevant databases to work on, however, they differ significantly on the sound durations, loudnesses normalization, categories, evaluations, etc.

\subsection{Machine learning-based prediction methodology}

A central obstacle shared by all predictive approaches is the scarcity of viable data augmentation strategies. Modifying the spectral or temporal properties of an audio signal can alter its perceived valence and arousal. A study~\cite{Wilkie2020} evaluated the effect of reverberation and amplitude on affective judgments across three sound sources. The result suggests that the perceptual impact of acoustic transformations could inform augmentation pipelines, though a more systematic mapping of transformation-to-affect relationships would be needed before any augmentation strategy could be applied reliably.

Prior work on affective audio prediction also points to recurring limitations in generalization. Classical approaches such as Support Vector Regression (SVR)~\cite{Abri} or feature-selection methods~\cite{Rey2025} rely on hand-crafted acoustic features that tend to transfer poorly across datasets.  Deep learning approaches reduce this dependency: convolutional and recurrent architectures trained directly on spectrograms have been applied to dimensional soundscape and audio emotion recognition~\cite{Ntalampiras2018, Serradilla2025}, but such models typically require large amounts of labeled data, a condition often unmet in affective audio corpora. To compensate, prior work segments each 6-second EmoSoundscapes clip into 1-second chunks. When these chunks overlap, \cite{Serradilla2025} shows that the test data are contaminated and reported scores become overly optimistic. Even without overlap, assigning labels to 1-second segments rests on an assumption of affective uniformity within each clip which was validated only at the 6-second scale~\cite{Fan2017} and which we do not expect to hold given our objectives. We did not re-evaluate these models, but we suspect these chunk-level scores overfit the dataset rather than reflecting robust generalization.


Our approach must generalize from a small number of evaluated examples. Furthermore, evaluation metrics should reflect the structure of the affective space, yet it is not clear that ($R^2$, RMSE) are well suited to capturing it.

The following section presents our approach, which adapts a pre-trained audio–text foundation model to Russell’s circumplex model of affect using low-rank fine-tuning, and introduces evaluation metrics that assess the structure of the valence–arousal space of stimuli. The model is trained on the AAD dataset, while EmoSoundscapes is reserved as a held-out cross-domain test set.

\section{Methodology and proposed solution}

\subsection{Bimodal text-audio foundation model}
Affective computing aims to endow machines with the ability to recognize and represent emotion~\cite{picard1997affective}; we approach this through a bimodal audio--text model. Foundation models pre-trained on audio--text data encode broad acoustic and semantic structure, a critical advantage when labeled affective data are scarce. A bimodal architecture is motivated by established links between language and affective perception. Sound identification has been shown to directly modulate perceived pleasantness \cite{heller2022}, showing that semantic categorization and affective judgement interact. Moreover, language functions primarily as a medium for transmitting internal states \cite{Fedorenko2024}, and the emotional lexicon constitutes a structured, socially shared representation of affective space \cite{Poria2017}. Contrastive text--audio training leverages this property by constraining audio embeddings to align with the semantics and syntax encoded in language, a mechanism that has been shown to improve generalization over unimodal approaches~\cite{Poria2017,Elizalde2022,Chen2020} and enables zero-shot classification of unseen data~\cite{Xie2021}. Preliminary work by~\cite{belaref2026datadriven} shows that such embeddings can even recover the topological structure of Russell's circumplex.

We build on CLAP \cite{Wu2024}, a bimodal foundation model trained on $4{,}325$ hours of audio--text pairs via contrastive learning \cite{Chen2020}. A central design choice is to avoid a regression head mapping CLAP embeddings directly to $(V,A)$ coordinates. Instead, we represent Russell's circumplex using 36 text anchors — eight axis labels and 28 affective descriptors \cite{Russell1980} — within the embedding space, and score audio $(V,A)$ by cosine similarity to these textual anchors.

All trainable parameters are adapted via LoRA ($r{=}64$)~\cite{Houlsby2019,Hu2021}, reducing overfitting on the small AAD training set. Fine-tuning proceeds in two phases: Phase~1 aligns the text encoder using the 36 subjective-feeling anchors of Fig.~\ref{fig:russell}, and Phase~2 fine-tunes the audio encoder to project audio embeddings towards their nearest text anchors.

\subsection{Dataset and split strategy}
\begin{figure}[t]
    \centering
    \includegraphics[width=7.25cm]{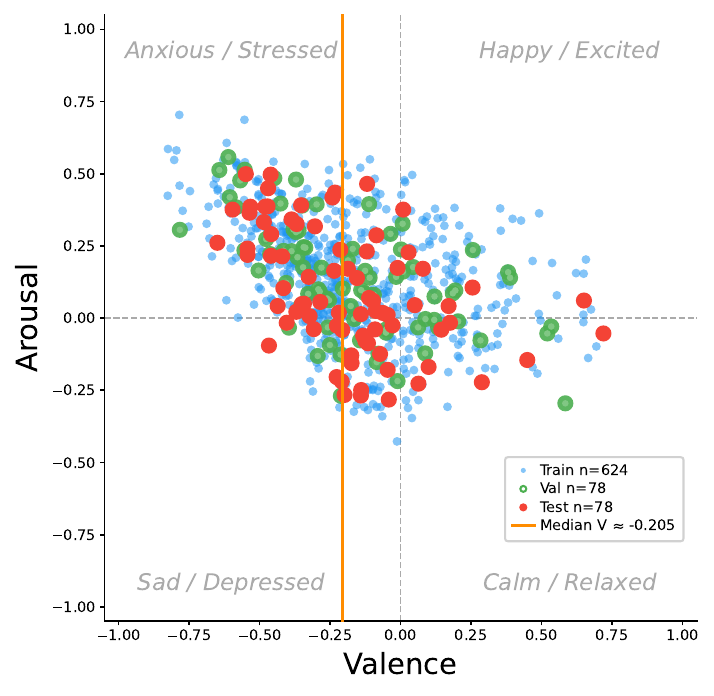}
    \caption{Distribution of AAD sounds in the $(V,A)$ space
    (train / val / test splits) stratified by valence median.}
    \label{fig:dataset}
\end{figure}

The proposed AAD split (624/78/78 train/val/test) was stratified by median valence ($V{=}{-}0.205$, seed = 42), with validation and test samples drawn proportionally from each stratum (Fig.~\ref{fig:dataset}).
Although AAD draws on the BBC sound-effect library, which is part of CLAP's
training set, CLAP never saw the segmented, normalized, and affect-labeled
versions used here. EmoSoundscapes \cite{Fan2017} ($n{=}1{,}213$) was used as a held-out external test set. 

\subsection{Anchor lexicon construction}
Since prediction relies on the cosine similarity between audio embeddings and each anchor's text prompt, the choice of prompt determines the anchor's placement in the embedding space. For each anchor, a reference audio embedding ($n{=}36$) was selected from the AAD training set by identifying the sound whose valence–arousal labels were closest to the corresponding position in Russell’s circumplex model.

This reference embedding was then used to select the most suitable text prompt from a pool of 728 templates and words (filtered from 1{,}005 entries by removing non-relevant terms \cite{demszky2020,cavanaugh2016}): each prompt was scored by its cosine similarity to the audio anchor's embedding, and the closest prompt was retained. Axis anchors were assigned first, and a global duplication constraint prevented two anchors from receiving the same prompt. To verify that each selected prompt occupied the correct region of the affective plane, its position in ($V,A$) space was estimated by comparing its text embedding to audio recordings that ground the two Russell axes. The \emph{audio-driven cosine} text selection yielded a Voronoi accuracy of $94.4\%$ on the Russell circumplex, defined as the fraction of anchors assigned to their correct Voronoi region.

\subsection{Fine-tuning strategy}

\begin{figure}[t]
    \centering
    \includegraphics[width=8.2cm]{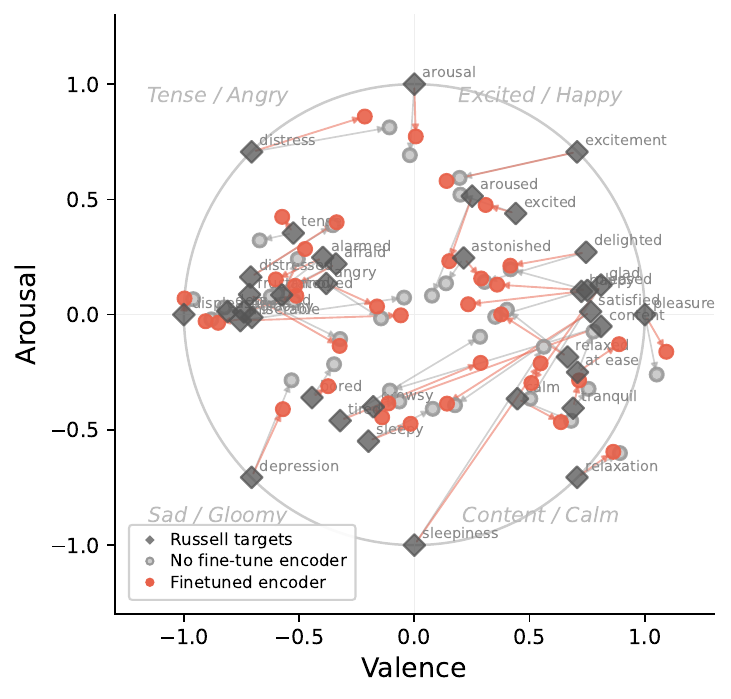}
    \caption{PCA of text anchor embeddings before and after
    text encoder fine-tuning. Voronoi accuracy improves from
    $94.4\%$ to $97.2\%$.}
    \label{fig:text_encoder_pca}
\end{figure}

\textbf{Phase 1 — Text Compass.}
The text encoder was fine-tuned while the audio encoder was frozen. We used the Manifold Kernel Alignment loss (MKA, \cite{Islam2025}), a manifold-aware variant of kernel alignment \cite{Wang2015}. It aligned the 36 anchor embeddings with their target positions in Russell's circumplex, without using audio waveforms or annotated audio–text pairs in the training objective. A lightweight TextMLP ($\text{Linear}(512{,}512){\to}\text{Dropout}(0.1){\to}\text{ReLU}{\to}\text{Linear}(512{,}512)$) further improved anchor placement (Fig.~\ref{fig:text_encoder_pca}).

\textbf{Phase 2 — Audio Alignment.}
To fine-tune the audio encoder, the Phase~1 text encoder was frozen, and a lightweight AudioMLP of the same architecture was appended to the audio encoder output. We explored three loss components:
\begin{description}
  \item[$\mathcal{L}_\text{anc}$] \emph{Anchor loss}: MSE between
    predicted $(V,A)$ and true labels local; per-sample supervision.
  \item[$\mathcal{L}_{KL}$] \textit{Symmetric Kullback--Leibler divergence}: supervision   is distributed across all 36 anchors as a soft target weighted by proximity to the true  $(V,A)$ label, following the use of a KL objective to fine-tune CLAP for affective prediction in~\cite{Pan2023}.      
  \item[$\mathcal{L}_\text{MKA}$] \emph{MKA audio loss} \cite{Islam2025}: aligns the
    global geometry of the audio embedding space with the $(V,A)$ label structure, without per-sample supervision.
\end{description}

The total loss combines up to three components:
\begin{equation}
  \mathcal{L} = \lambda_1\,\mathcal{L}_\text{anc}
              + \lambda_2\,\mathcal{L}_\text{KL}
              + \lambda_3\,\mathcal{L}_\text{MKA}
\end{equation}

Ablation studies allowed us to test various weighting ($\lambda_1, \lambda_2, \lambda_3 \in \{0, 0.25, 0.5, 1\}$) and key architectural choices.
All models were trained with batch size $n{=}32$.
Four configurations isolated the contribution of each stage:
\textit{Audio-only} fine-tuned only the audio encoder ($\mathcal{L}_\text{anc}$+$\mathcal{L}_\text{KL}{\times}0.5$);
\textit{Audio-Text} added Phase~1 text pre-alignment ($\mathcal{L}_\text{anc}$+$\mathcal{L}_\text{KL}{\times}0.25$);
\textit{Full-Pipeline} re-tuned the text encoder after Phase~2 ($\mathcal{L}_\text{anc}$+$\mathcal{L}_\text{MKA}$);
\textit{Full-Pipeline\,+\,KL} added a KL term ($+\mathcal{L}_\text{KL}{\times}0.25$).

\subsection{Evaluation} 
 Beyond standard metrics ($R^2$, RMSE), and given variability in labeling protocols across datasets, we assessed model robustness using additional evaluations. We selected metrics designed to capture the topology of the embedding space, as well as the relative ordering of sound ratings along each axis (arousal and valence).

\begin{description}
    \item[Sp$\uparrow$] \emph{Spearman rank correlation}~\cite{Spearman1904} : measures whether the predicted ordering of sounds along each affective axis matches the true ordering, regardless of absolute scale. Used as primary selection criterion on EmoSoundscapes (out-of-distribution test).
    
    \item[RSA$\uparrow$] \emph{Representational Similarity Analysis}~\cite{Kriegeskorte2008} : measures whether the pairwise distance structure of the 512-dimensional audio embedding space reflects the 2D $(V,A)$ label space, via Spearman correlation of the two distance matrices.
    
    \item[Trust$\uparrow$] \emph{Trustworthiness}~\cite{Venna2001} : checks whether neighbors in the embedding are also neighbors in the $(V,A)$ space.
    
    \item[$\theta$-Sp$\uparrow$] \emph{Angular Spearman}: measures whether the circular order of emotions on Russell's circle is preserved: $\theta = \arctan2(\hat{A}, \hat{V})$.
    
    \item[Proc$\downarrow$]  \emph{Procrustes disparity}~\cite{Gower1975} : aligns the predicted $(V, A)$ point cloud to the true labels via optimal rotation, scaling, and translation, then measures the residual distance; it captures global shape preservation independently of per-axis calibration, complementing $R^2$.
\end{description}

\subsection{Results}
\textit{Audio-only} already yielded a large gain ($\text{Sp}(V){:}\ {-}0.09\!\to\!0.71$ on AAD), showing that fine-tuning only the audio encoder generalized (Table~\ref{tab:ablation}). Adding Phase~1 \textit{Audio-Text} brought a consistent but modest further improvement across 
datasets (Fig.~\ref{fig:text_encoder_pca}). Re-tuning the text encoder after Phase~2 \textit{Full-Pipeline} produced the largest cross-domain gain on EmoSoundscapes ($\text{Sp}(V){=}0.705$) and achieved the best Procrustes disparity ($0.514$), indicating that the 
affective geometry is better preserved. Including the $\mathcal{L}_\text{KL}$ loss \textit{Full-Pipeline\,+\,KL} instead maximized representation: it reached the highest RSA ($0.353$), Trustworthiness ($0.674$), and Sp(A) across both datasets. Across all fine-tuned models, RSA significance was confirmed by a Mantel permutation test ($p{<}0.001$)~\cite{Mantel1967}.

\begin{table*}[!b]
\centering
\caption{Ablation on EmoSoundscapes ($n{=}1{,}213$) and AAD test ($n{=}78$).}
\label{tab:ablation}
\small
\begin{tabular}{|l|cccccc|cccccc|}
\hline
 & \multicolumn{6}{c|}{\textbf{EmoSoundscapes ($n=1{,}213$)}}
 & \multicolumn{6}{c|}{\textbf{AAD test ($n=78$)}} \\
\textbf{Model}
  & Sp(V)$\uparrow$ & Sp(A)$\uparrow$ & RSA$\uparrow$ & Trust$\uparrow$ & Proc$\downarrow$ & $\theta$-Sp$\uparrow$
  & Sp(V)$\uparrow$ & Sp(A)$\uparrow$ & R$^2$(V)$\uparrow$ & R$^2$(A)$\uparrow$ & Trust$\uparrow$ & Proc$\downarrow$ \\
\hline
Zero-shot CLAP
  & 0.622 & 0.226 & 0.278 & 0.645 & 0.697 & 0.338
  & $-$0.094 & 0.234 & $-$2.762 & $-$1.111 & 0.588 & 0.955 \\
Audio-only \ 
  & 0.647 & 0.684 & 0.322 & 0.665 & 0.586 & 0.501
  & 0.714 & 0.857 & 0.605 & 0.680 & 0.631 & 0.327 \\
Audio-Text
  & 0.657 & 0.681 & 0.327 & 0.666 & 0.569  & \textbf{0.517}
  & 0.724 & \textbf{0.854} & 0.616 & 0.684 & 0.630 & 0.315 \\
Full-Pipeline
  & \textbf{0.705} & 0.687 & 0.347 & 0.671 & \textbf{0.514} & 0.506
  & \textbf{0.731} & 0.829 & 0.608 & \textbf{0.696} & 0.637 & \textbf{0.327} \\
Full-Pipeline\,+\,KL
  & 0.663 & \textbf{0.705} & \textbf{0.353} & \textbf{0.674} & 0.546 & 0.486
  & 0.722 & 0.853 & \textbf{0.625} & 0.690 & \textbf{0.640} & 0.314 \\
\hline
\end{tabular}
\end{table*}

\section{Future work and expected contributions to Affective Computing}

\subsection{Open Issues/discussion}
The methodology presented above shows promising generalization performance, but several directions remain open. On the modeling side, only a limited portion of the fine-tuning space has been explored; alternative contrastive losses and adaptation strategies are still to be evaluated. Beyond the encoder outputs considered here, we have begun investigating the fine-tuning of other components of the audio encoder, in particular the attention layers of its Swin-Transformer backbone (hierarchical token-semantic audio transformer HTS-AT~\cite{Chen2022}), which controls how the model attends to regions of the input spectrogram. Adapting this attention mechanism could allow the model to place greater emphasis on acoustic regions that are more relevant to affective perception.


A more fundamental question concerns transferability to atypical listening conditions. Because CLAP already encodes a strong prior over affective organization, evident even in zero-shot evaluations, it remains unclear whether the model can faithfully reorganize its representation around labels produced by listeners whose affective responses differ 
from the neurotypical norm, or whether this prior will dominate and mask individual-specific structure. This tension between a useful pretrained prior and the need for genuine personalization is a central open problem for the remainder of this thesis.

We also note two scope restrictions: spatialization is excluded, as we focus on monophonic signals. Finally, AAD provides only aggregated labels and no measure of inter-rater variability, disparity between individual evaluations may itself constitute an informative signal. Modeling this variability could help characterize the heterogeneity reported in atypical auditory processing~\cite{Poulsen2024}.



\subsection{Probing the models}

A final direction concerns probing the trained models to identify which acoustic features drive affective predictions. Reverse-correlation and bubble-based methods have been used to expose the spectro-temporal cues underlying machine-learning classifiers \cite{Thoret2021} and could be applied here to reveal which components of a sound influence predicted valence and arousal. The complementary question -- how controlled acoustic transformations affect perceived emotion \cite{Wilkie2020} -- would inform label-aware data augmentation strategies, directly addressing the augmentation problem raised in Section~\ref{sec:2}.

\newpage{}
\section*{Ethical impact statement}

This work seeks a framework that learns affective representations in a resource-efficient way: by generalizing from a small number of labeled examples, it reduces the annotation workload and enables lightweight, low-footprint models. The present study relies exclusively on publicly available datasets collected from neurotypical participants; no human data were collected here. Our longer-term objective is to model affective responses for autistic listeners with auditory hypersensitivity, a vulnerable population, and we have begun a collaboration with a specialized autism resource center (CRAIF) toward this goal. No data collection involving human participants will be conducted without approval by a French ethic committee.

\bibliographystyle{IEEEtran}

\bibliography{ref}

\begin{thebibliography}{10}
\providecommand{\url}[1]{#1}
\csname url@samestyle\endcsname
\providecommand{\newblock}{\relax}
\providecommand{\bibinfo}[2]{#2}
\providecommand{\BIBentrySTDinterwordspacing}{\spaceskip=0pt\relax}
\providecommand{\BIBentryALTinterwordstretchfactor}{4}
\providecommand{\BIBentryALTinterwordspacing}{\spaceskip=\fontdimen2\font plus
\BIBentryALTinterwordstretchfactor\fontdimen3\font minus \fontdimen4\font\relax}
\providecommand{\BIBforeignlanguage}[2]{{%
\expandafter\ifx\csname l@#1\endcsname\relax
\typeout{** WARNING: IEEEtran.bst: No hyphenation pattern has been}%
\typeout{** loaded for the language `#1'. Using the pattern for}%
\typeout{** the default language instead.}%
\else
\language=\csname l@#1\endcsname
\fi
#2}}
\providecommand{\BIBdecl}{\relax}
\BIBdecl

\bibitem{Poulsen2024}
R.~Poulsen, Z.~Williams, P.~Dwyer, E.~Pellicano, P.~F. Sowman, and D.~McAlpine, ``How auditory processing influences the autistic profile: A review,'' \emph{Autism Res.}, vol.~17, no.~12, pp. 2452--2470, Dec. 2024.

\bibitem{Kanakri2017}
S.~M. Kanakri, M.~Shepley, J.~W. Varni, and L.~G. Tassinary, ``Noise and autism spectrum disorder in children: An exploratory survey,'' \emph{Res. Dev. Disabil.}, vol.~63, pp. 85--94, 2017.

\bibitem{Landowska2022}
A.~Landowska, A.~Karpus, T.~Zawadzka, B.~Robins \emph{et~al.}, ``Automatic emotion recognition in children with autism: A systematic literature review,'' \emph{Sensors}, vol.~22, no.~4, p. 1649, 2022.

\bibitem{OConnor2012}
K.~O'Connor, ``Auditory processing in autism spectrum disorder: A review,'' \emph{Neurosci. Biobehav. Rev.}, vol.~36, no.~2, pp. 836--854, 2012.

\bibitem{Kwong2025}
T.~C. Kwong, H.-L. Yuan, S.~W.~Y. Mung, H.~K. Chu \emph{et~al.}, ``Intervention technology of aural perception controllable headset for children with autism spectrum disorder,'' \emph{Sci. Rep.}, vol.~15, no.~1, p. 5356, 2025.

\bibitem{fang2025metaperser}
S.-X. Fang, L.-Y. Shen, Y.-C. Lin, H.-C. Chou, and H.-y. Lee, ``Meta-{PerSER}: Few-shot listener personalized speech emotion recognition via meta-learning,'' in \emph{Proc. Interspeech}, 2025, pp. 1--5.

\bibitem{Russell1980}
J.~A. Russell, ``A circumplex model of affect,'' \emph{J. Pers. Soc. Psychol.}, vol.~39, no.~6, pp. 1161--1178, 1980.

\bibitem{Fan2017}
J.~Fan, M.~Thorogood, and P.~Pasquier, ``{Emo-Soundscapes}: A dataset for soundscape emotion recognition,'' in \emph{Proc. 7th Int. Conf. Affective Comput. Intell. Interaction (ACII)}, 2017, pp. 196--201.

\bibitem{Yang2018}
W.~Yang, K.~Makita, T.~Nakao, N.~Kanayama \emph{et~al.}, ``Affective auditory stimulus database: An expanded version of the international affective digitized sounds ({IADS-E}),'' \emph{Behav. Res. Methods}, vol.~50, no.~4, pp. 1415--1429, 2018.

\bibitem{IADS2}
M.~M. Bradley and P.~J. Lang, ``The {International Affective Digitized Sounds (IADS-2)}: Affective ratings of sounds and instruction manual,'' University of Florida, Gainesville, FL, USA, Tech. Rep. B-3, 2007.

\bibitem{AAD}
H.~Ridley, S.~Cunningham, J.~Darby, J.~Henry, and R.~Stocker, ``The affective audio dataset ({AAD}) for non-musical, non-vocalized, audio emotion research,'' \emph{IEEE Trans. Affect. Comput.}, vol.~16, no.~1, pp. 394--404, Jan. 2025.

\bibitem{Wilkie2020}
S.~Wilkie and T.~Stockman, ``The effect of audio cues and sound source stimuli on the perception of approaching objects,'' \emph{Appl. Acoust.}, vol. 167, p. 107388, Oct. 2020.

\bibitem{Abri}
F.~Abri, L.~F. Guti{\'e}rrez, A.~S. Namin, D.~R.~W. Sears, and K.~S. Jones, ``Predicting emotions perceived from sounds,'' in \emph{Proc. IEEE Int. Conf. Big Data}, Atlanta, GA, USA, 2020, pp. 2057--2064.

\bibitem{Rey2025}
S.~Rey, L.~Martino, R.~S. Mill{\'a}n, and E.~Morgado, ``Feature selection via graph topology inference for soundscape emotion recognition,'' \emph{arXiv preprint arXiv:2509.16760}, 2025.

\bibitem{Ntalampiras2018}
S.~Ntalampiras, ``Soundscape emotion recognition via deep learning,'' in \emph{Proc. Int. Conf. Affective Comput. Intell. Interaction (ACII)}, 2018.

\bibitem{Serradilla2025}
F.~Serradilla, {\'A}.~San~Juan, and D.~Mart{\'i}nez-I{\~n}igo, ``Emotional parameter estimation from {Emo-Soundscapes} dataset using deep convolutional autoencoders,'' \emph{Multimedia Tools Appl.}, vol.~84, no.~24, pp. 28\,693--28\,707, 2025.

\bibitem{picard1997affective}
R.~W. Picard, \emph{Affective Computing}.\hskip 1em plus 0.5em minus 0.4em\relax Cambridge, MA, USA: MIT Press, 1997.

\bibitem{heller2022}
L.~M. Heller and J.~M. Smith, ``Identification of everyday sounds affects their pleasantness,'' \emph{Front. Psychol.}, vol.~13, Jul. 2022.

\bibitem{Fedorenko2024}
E.~Fedorenko, S.~T. Piantadosi, and E.~A.~F. Gibson, ``Language is primarily a tool for communication rather than thought,'' \emph{Nature}, vol. 630, no. 8017, pp. 575--586, 2024.

\bibitem{Poria2017}
S.~Poria, E.~Cambria, R.~Bajpai, and A.~Hussain, ``A review of affective computing: From unimodal analysis to multimodal fusion,'' \emph{Inf. Fusion}, vol.~37, pp. 98--125, 2017.

\bibitem{Elizalde2022}
B.~Elizalde, S.~Deshmukh, M.~A. Ismail, and H.~Wang, ``{CLAP}: Learning audio concepts from natural language supervision,'' \emph{arXiv preprint arXiv:2206.04769}, 2022.

\bibitem{Chen2020}
T.~Chen, S.~Kornblith, M.~Norouzi, and G.~Hinton, ``A simple framework for contrastive learning of visual representations,'' in \emph{Proc. 37th Int. Conf. Mach. Learn. (ICML)}, 2020, pp. 1597--1607.

\bibitem{Xie2021}
H.~Xie and T.~Virtanen, ``Zero-shot audio classification via semantic embeddings,'' \emph{IEEE/ACM Trans. Audio, Speech, Lang. Process.}, vol.~29, pp. 1233--1242, 2021.

\bibitem{belaref2026datadriven}
A.~Belaref, S.~Sadok, Z.~Noumir, and R.~S\'eguier, ``Data-driven decoding of {Russell}'s circumplex model of affect,'' \emph{arXiv preprint arXiv:2606.16843}, 2026.

\bibitem{Wu2024}
Y.~Wu, K.~Chen, T.~Zhang, Y.~Hui, T.~Berg-Kirkpatrick, and S.~Dubnov, ``Large-scale contrastive language-audio pretraining with feature fusion and keyword-to-caption augmentation,'' in \emph{Proc. IEEE Int. Conf. Acoust., Speech, Signal Process. (ICASSP)}, 2023, pp. 1--5.

\bibitem{Houlsby2019}
N.~Houlsby, A.~Giurgiu, S.~Jastrzebski, B.~Morrone \emph{et~al.}, ``Parameter-efficient transfer learning for {NLP},'' in \emph{Proc. 36th Int. Conf. Mach. Learn. (ICML)}, vol.~97, 2019, pp. 2790--2799.

\bibitem{Hu2021}
E.~J. Hu, Y.~Shen, P.~Wallis, Z.~Allen-Zhu, Y.~Li, S.~Wang, L.~Wang, and W.~Chen, ``{LoRA}: Low-rank adaptation of large language models,'' in \emph{Proc. Int. Conf. Learn. Represent. (ICLR)}, 2022.

\bibitem{demszky2020}
D.~Demszky, D.~Movshovitz-Attias, J.~Ko, A.~Cowen, G.~Nemade, and S.~Ravi, ``{GoEmotions}: A dataset of fine-grained emotions,'' in \emph{Proc. 58th Annu. Meeting Assoc. Comput. Linguistics}, Online, 2020, pp. 4040--4054.

\bibitem{cavanaugh2016}
L.~A. Cavanaugh, D.~J. MacInnis, and A.~M. Weiss, ``Perceptual dimensions differentiate emotions,'' \emph{Cogn. Emot.}, vol.~30, no.~8, pp. 1430--1445, Dec. 2016.

\bibitem{Islam2025}
M.~T. Islam, D.~Liu, and D.~Sarkar, ``Manifold approximation leads to robust kernel alignment,'' \emph{arXiv preprint arXiv:2510.22953}, 2025.

\bibitem{Wang2015}
T.~Wang, D.~Zhao, and S.~Tian, ``An overview of kernel alignment and its applications,'' \emph{Artif. Intell. Rev.}, vol.~43, no.~2, pp. 179--192, 2015.

\bibitem{Pan2023}
Y.~Pan, Y.~Hu, Y.~Yang, W.~Fei, J.~Yao, H.~Lu, L.~Ma, and J.~Zhao, ``{GEmo-CLAP}: Gender-attribute-enhanced contrastive language-audio pretraining for accurate speech emotion recognition,'' \emph{arXiv preprint arXiv:2306.07848}, 2023.

\bibitem{Spearman1904}
C.~Spearman, ``The proof and measurement of association between two things,'' \emph{Amer. J. Psychol.}, vol.~15, no.~1, pp. 72--101, 1904.

\bibitem{Kriegeskorte2008}
N.~Kriegeskorte, M.~Mur, and P.~A. Bandettini, ``Representational similarity analysis --- connecting the branches of systems neuroscience,'' \emph{Front. Syst. Neurosci.}, vol.~2, p.~4, 2008.

\bibitem{Venna2001}
J.~Venna and S.~Kaski, ``Neighborhood preservation in nonlinear projection methods: An experimental study,'' in \emph{Proc. Int. Conf. Artif. Neural Netw. (ICANN)}.\hskip 1em plus 0.5em minus 0.4em\relax Springer, 2001, pp. 485--491.

\bibitem{Gower1975}
J.~C. Gower, ``Generalized {Procrustes} analysis,'' \emph{Psychometrika}, vol.~40, no.~1, pp. 33--51, 1975.

\bibitem{Mantel1967}
N.~Mantel, ``The detection of disease clustering and a generalized regression approach,'' \emph{Cancer Res.}, vol.~27, no.~2, pp. 209--220, 1967.

\bibitem{Chen2022}
K.~Chen, X.~Du, B.~Zhu, Z.~Ma, T.~Berg-Kirkpatrick, and S.~Dubnov, ``{HTS-AT}: A hierarchical token-semantic audio transformer for sound classification and detection,'' in \emph{Proc. IEEE Int. Conf. Acoust., Speech, Signal Process. (ICASSP)}, 2022, pp. 646--650.

\bibitem{Thoret2021}
E.~Thoret, T.~Andrillon, D.~L{\'e}ger, and D.~Pressnitzer, ``Probing machine-learning classifiers using noise, bubbles, and reverse correlation,'' \emph{J. Neurosci. Methods}, vol. 362, p. 109297, Oct. 2021.

\end{thebibliography}

\end{document}